\PassOptionsToPackage{unicode}{hyperref}
\PassOptionsToPackage{hyphens}{url}
\PassOptionsToPackage{dvipsnames,svgnames,x11names}{xcolor}
\documentclass[
  11pt,
]{article}
\usepackage{amsmath,amssymb}
\usepackage{iftex}
\ifPDFTeX
  \usepackage[T1]{fontenc}
  \usepackage[utf8]{inputenc}
  \usepackage{textcomp} 
\else 
  \usepackage{unicode-math} 
  \defaultfontfeatures{Scale=MatchLowercase}
  \defaultfontfeatures[\rmfamily]{Ligatures=TeX,Scale=1}
\fi
\usepackage{lmodern}
\ifPDFTeX\else
\fi
\IfFileExists{upquote.sty}{\usepackage{upquote}}{}
\IfFileExists{microtype.sty}{
  \usepackage[]{microtype}
  \UseMicrotypeSet[protrusion]{basicmath} 
}{}
\makeatletter
\@ifundefined{KOMAClassName}{
  \IfFileExists{parskip.sty}{%
    \usepackage{parskip}
  }{
    \setlength{\parindent}{0pt}
    \setlength{\parskip}{6pt plus 2pt minus 1pt}}
}{
  \KOMAoptions{parskip=half}}
\makeatother
\usepackage{xcolor}
\usepackage[margin=1in]{geometry}
\usepackage{longtable,booktabs,array}
\usepackage{calc} 
\usepackage{etoolbox}
\makeatletter
\patchcmd\longtable{\par}{\if@noskipsec\mbox{}\fi\par}{}{}
\makeatother
\IfFileExists{footnotehyper.sty}{\usepackage{footnotehyper}}{\usepackage{footnote}}
\makesavenoteenv{longtable}
\usepackage{graphicx}
\makeatletter
\def\maxwidth{\ifdim\Gin@nat@width>\linewidth\linewidth\else\Gin@nat@width\fi}
\def\maxheight{\ifdim\Gin@nat@height>\textheight\textheight\else\Gin@nat@height\fi}
\makeatother
\setkeys{Gin}{width=\maxwidth,height=\maxheight,keepaspectratio}
\makeatletter
\def\fps@figure{htbp}
\makeatother
\providecommand{\tightlist}{%
  \setlength{\itemsep}{0pt}\setlength{\parskip}{0pt}}
\ifLuaTeX
  \usepackage{selnolig}  
\fi
\IfFileExists{bookmark.sty}{\usepackage{bookmark}}{\usepackage{hyperref}}
\IfFileExists{xurl.sty}{\usepackage{xurl}}{} 
\hypersetup{
  pdftitle={Conversational Domain Adaptation of IndicTrans2 across 21 Indic Languages via Experience Replay and Model Soups},
  pdfauthor={Aditya Pratap Singh},
  colorlinks=true,
  linkcolor={blue},
  filecolor={Maroon},
  citecolor={Blue},
  urlcolor={blue},
  pdfcreator={LaTeX via pandoc}}

\title{Conversational Domain Adaptation of IndicTrans2 across 21 Indic
Languages via Experience Replay and Model Soups}
\author{Aditya Pratap Singh\\ \texttt{adipras1407@gmail.com}}
\date{June 2026}

\begin{document}
\maketitle

\hypertarget{abstract}{%
\subsection{Abstract}\label{abstract}}

IndicTrans2 is the strongest open English to Indic translation system,
but like most systems it is trained on general text and tends to sound
stiff on casual, conversational input. We adapt IndicTrans2-1B to
conversational register across all 21 Indic languages using only public
data (OpenSubtitles, BPCC-H-Daily, Tatoeba). Plain fine-tuning improves
conversational chrF but forgets the general domain (it drops 3.9 chrF on
FLORES for Hindi). Mixing general data back into training (experience
replay) and then averaging the fine-tuned weights with the base (model
souping) removes that trade-off: the resulting model beats
IndicTrans2-1B on conversational chrF in every one of the 21 languages
(mean +6.2) while matching it on FLORES (mean change -0.17, all within
0.7 chrF). Paired bootstrap tests confirm the conversational gains are
significant (p $\le$ 0.004) and that FLORES is not significantly degraded.
We are deliberate about scope: these are chrF gains, and a blind human
plus multi-model LLM check does not confirm them as a perceived quality
improvement, so we treat the conversational gain as largely a register
match to the references rather than proof of better translation. The
techniques are not new; the contribution is the honest, end-to-end study
in the Indic conversational setting.

\hypertarget{introduction}{%
\subsection{1. Introduction}\label{introduction}}

Most translation systems are tuned and measured on general benchmarks
like FLORES-200. But a lot of real translation is conversational: chat,
subtitles, voice commands. That text is shorter, more idiomatic, and
more elliptical than news, and a model that scores well on FLORES can
still produce stilted output on a casual line.

For Indian languages, IndicTrans2 is the best open option, trained on
the 230M-pair BPCC corpus. It is a general-domain model, so we asked a
simple question:

\begin{quote}
Can we adapt a strong multilingual model to conversational register
across many languages, without hurting its general quality, using only
public data?
\end{quote}

By chrF the answer looks like a clear yes. Plain fine-tuning causes the
expected catastrophic forgetting, and experience replay plus model
souping fixes it while delivering large conversational gains across all
21 languages with no significant FLORES loss (§5).

We did not stop there. The whole point is a claim about translation
\emph{quality}, not about chrF, so we also ran a blind human and LLM
evaluation (§6). That check does not confirm a quality gain. We report
both results honestly: the recipe works by the metric, but the metric
overstates how much it helps in conversation.

\textbf{Contributions.}

\begin{enumerate}
\def\labelenumi{\arabic{enumi}.}
\tightlist
\item
  A reproducible study of conversational adaptation for English to Indic
  translation across 21 languages, built entirely from public data, with
  a working anti-forgetting recipe and significance tests.
\item
  An honest account of the recipe's limits: forgetting under plain
  fine-tuning, inflated gains on easy in-domain test sets, a metric that
  overstates conversational quality, and a floor set by the base model
  on the lowest-resource languages.
\end{enumerate}

The techniques we use, model soups and replay, are established. The
value here is the systematic study and its honest evaluation in a
setting nobody had measured this way.

\hypertarget{related-work}{%
\subsection{2. Related Work}\label{related-work}}

\textbf{Domain adaptation for NMT.} Fine-tuning a general model on
in-domain data is standard, and its main failure mode is forgetting the
general domain. Mixing general data back in during fine-tuning is a
known fix and a form of experience replay {[}Chu et al., 2017; Robins,
1995{]}.

\textbf{Model soups.} Averaging the weights of fine-tuned models
improves accuracy {[}Wortsman et al., 2022a{]}; WiSE-FT averages a
fine-tuned model with its starting point to balance in-domain and
out-of-domain performance {[}Wortsman et al., 2022b{]}. We apply that
base-to-fine-tune averaging to translation.

\textbf{Metrics versus human judgment.} chrF and BLEU reward overlap
with the reference and can diverge from human assessment
{[}Callison-Burch et al., 2006; Mathur et al., 2020; Freitag et al.,
2021{]}. When adaptation shifts a model toward the reference style, the
metric can rise without quality rising. Our result is a concrete case of
this in conversational Indic translation.

\textbf{Indic MT.} IndicTrans2 {[}Gala et al., 2023{]} and BPCC are the
base model and a data source. WAT hosts Indic translation shared tasks
{[}Nakazawa et al., 2024{]}.

\hypertarget{data}{%
\subsection{3. Data}\label{data}}

\textbf{Base model.} \texttt{ai4bharat/indictrans2-en-indic-1B} (1.1B
parameters).

\textbf{Conversational data} (English to Indic, each pair tagged with
its target language):

\begin{itemize}
\tightlist
\item
  OPUS OpenSubtitles for the five Indic languages with substantial
  subtitle data (hi, ml, ta, te, ur), capped at 40k per language.
\item
  BPCC-H-Daily, human-written everyday and assistant utterances,
  available for all 21 languages (about 1.8k to 11k pairs each).
\item
  Tatoeba everyday sentences (Hindi).
\end{itemize}

After filtering for length, length ratio, exact duplicates, and script,
the conversational set is 294,119 pairs across 21 languages (train
283,753 / dev 4,183 / test 6,183). The Hindi-only runs also apply a
LaBSE cosine filter at 0.70 to drop misaligned subtitle pairs.

\textbf{General anchor} (for replay): BPCC-H-Wiki, a general register,
sampled roughly 1:1 with the conversational data (254,469 pairs). The
mixed training set is therefore 538,222 pairs.

Conversational volume is very uneven. The five OpenSubtitles languages
hold 30k to 48k pairs each; the other sixteen rely on BPCC-H-Daily
(about 2k to 11k). This skew is baked into the available data and it
shapes the results (§7).

\hypertarget{method}{%
\subsection{4. Method}\label{method}}

We compare three setups, all fine-tuning IndicTrans2-1B (IndicProcessor
preprocessing with per-example target-language tags, bf16, Adafactor,
single NVIDIA A10G GPU):

\begin{enumerate}
\def\labelenumi{\arabic{enumi}.}
\tightlist
\item
  \textbf{Plain FT.} Fine-tune on conversational data only.
\item
  \textbf{Mixed FT (replay).} Fine-tune on conversational plus the
  general anchor, about 1:1.
\item
  \textbf{Model soup.} Average the base and a fine-tuned model,
  \texttt{soup\ =\ (1\ -\ a)\ *\ base\ +\ a\ *\ finetuned}, sweeping the
  mixing weight \texttt{a} and picking the best
  conversational-to-general trade-off. We stream the two checkpoints'
  tensors so memory stays bounded.
\end{enumerate}

Hindi-only models use 146k conversational pairs (2 to 3 epochs). The
multilingual models use the full 294k conversational and 538k mixed sets
(2 epochs and 1 epoch, matched on examples seen).

We score chrF2 (sacreBLEU) on a held-out conversational test and on
FLORES-200 devtest, per language, with paired bootstrap significance for
representative languages.

\hypertarget{results}{%
\subsection{5. Results}\label{results}}

By the metric, the recipe works. These numbers are real; §6 then asks
whether they mean what they appear to.

\hypertarget{hindi-case-study}{%
\subsubsection{5.1 Hindi case study}\label{hindi-case-study}}

Table 1 isolates the mechanism on Hindi. Figure 1 shows the same story.

\textbf{Table 1: Hindi (chrF2).}

\begin{longtable}[]{@{}lll@{}}
\toprule\noalign{}
Model & conv & FLORES \\
\midrule\noalign{}
\endhead
\bottomrule\noalign{}
\endlastfoot
IndicTrans2 base & 51.6 & 61.7 \\
Plain conv-FT & \textbf{57.0} & 57.8 (forgetting) \\
Mixed FT & 56.0 & 60.6 \\
\textbf{Model soup (a=0.6)} & 54.1 & \textbf{61.8} \\
\end{longtable}

Plain fine-tuning gains 5.4 conversational chrF but loses 3.9 on FLORES,
which is clear forgetting. Mixed FT recovers most of the general
quality, and the soup lands at or above the base on both axes. Sweeping
\texttt{a} from 0.3 to 0.7 puts the knee at 0.6.

\begin{figure}
\centering
\includegraphics[width=0.78\textwidth,height=\textheight]{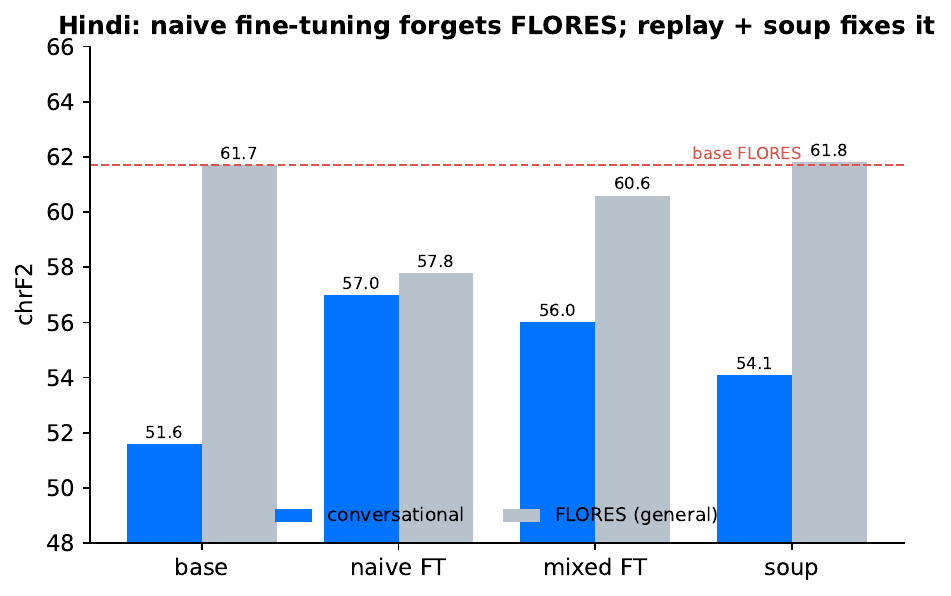}
\caption{On Hindi, plain fine-tuning lifts conversational chrF but drops
FLORES below the base. Replay keeps most of the general quality, and
averaging with the base restores FLORES while staying ahead on
conversation.}
\end{figure}

\hypertarget{all-21-languages}{%
\subsubsection{5.2 All 21 languages}\label{all-21-languages}}

The mixed soup improves conversational chrF in every language (mean
+6.2) and barely moves FLORES (mean -0.17, all within 0.7). Figure 2
shows the per-language gains; Figure 3 shows that the conversational
gain comes with essentially no change in general quality.

\textbf{Table 2: All 21 languages, base to mixed-soup (chrF2).}

\begin{longtable}[]{@{}lll@{}}
\toprule\noalign{}
Lang & conv (delta) & FLORES (delta) \\
\midrule\noalign{}
\endhead
\bottomrule\noalign{}
\endlastfoot
asm & 61.9 to 69.4 (+7.6) & 44.2 to 44.1 (-0.1) \\
ben & 68.7 to 73.6 (+4.8) & 58.1 to 58.1 (0.0) \\
brx & 62.1 to 68.0 (+5.9) & 45.9 to 46.0 (+0.1) \\
doi & 63.4 to 71.7 (+8.3) & 49.8 to 49.6 (-0.2) \\
gom & 69.4 to 75.6 (+6.2) & 49.1 to 49.2 (+0.1) \\
guj & 53.7 to 65.9 (+12.2) & 55.3 to 54.7 (-0.6) \\
hin & 53.9 to 56.8 (+2.9) & 60.0 to 59.6 (-0.5) \\
kan & 52.4 to 57.1 (+4.7) & 59.0 to 58.7 (-0.3) \\
kas & 45.8 to 51.3 (+5.5) & 39.1 to 38.8 (-0.4) \\
mai & 57.8 to 65.2 (+7.4) & 56.6 to 56.6 (0.0) \\
mal & 45.7 to 48.4 (+2.7) & 61.1 to 61.4 (+0.3) \\
mar & 70.2 to 77.3 (+7.1) & 55.0 to 54.5 (-0.5) \\
mni & 56.4 to 62.8 (+6.4) & 48.2 to 48.3 (+0.2) \\
npi & 68.8 to 72.1 (+3.3) & 59.4 to 59.0 (-0.5) \\
ory & 56.4 to 65.7 (+9.2) & 52.0 to 51.3 (-0.7) \\
pan & 68.0 to 76.5 (+8.4) & 55.3 to 54.8 (-0.5) \\
san & 54.5 to 63.6 (+9.1) & 36.3 to 36.5 (+0.2) \\
sat & 50.7 to 56.2 (+5.6) & 30.7 to 30.3 (-0.4) \\
tam & 52.7 to 55.1 (+2.4) & 64.1 to 64.4 (+0.3) \\
tel & 49.9 to 60.0 (+10.1) & 62.4 to 62.0 (-0.4) \\
urd & 48.1 to 49.4 (+1.3) & 54.7 to 54.7 (+0.1) \\
\textbf{mean} & \textbf{+6.2} & \textbf{-0.17} \\
\end{longtable}

\begin{figure}
\centering
\includegraphics[width=0.72\textwidth,height=\textheight]{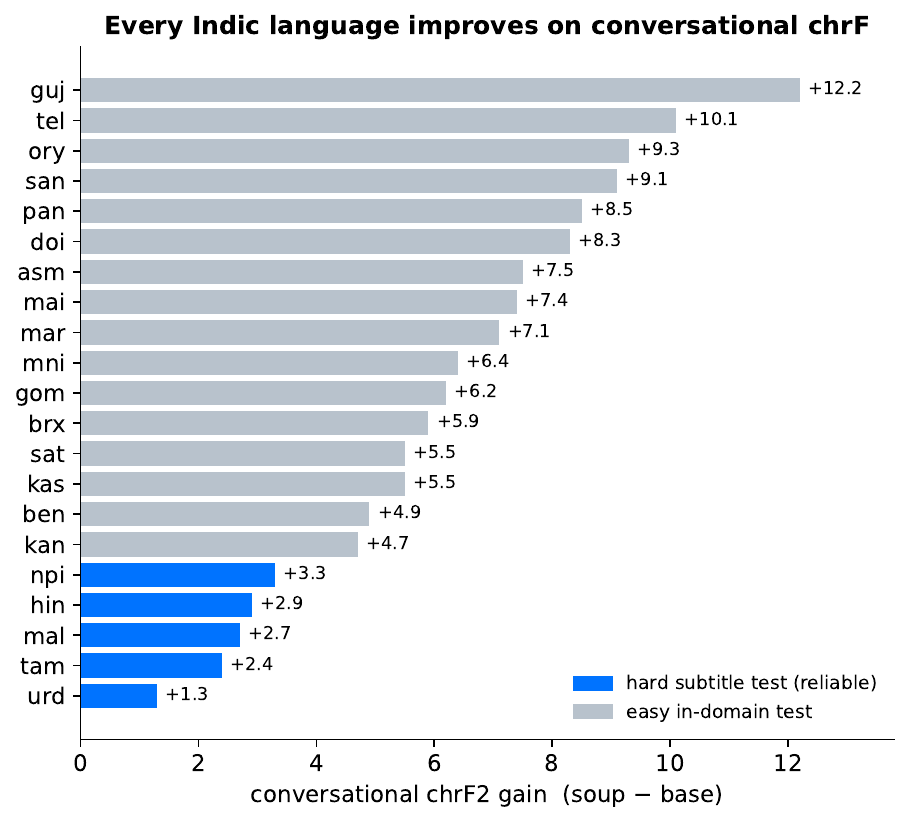}
\caption{Conversational chrF gain for all 21 languages. Blue marks the
five languages with a hard subtitle test set; these are the gains to
trust. The large grey gains come from easy in-domain test sets (§7).}
\end{figure}

\begin{figure}
\centering
\includegraphics[width=0.72\textwidth,height=\textheight]{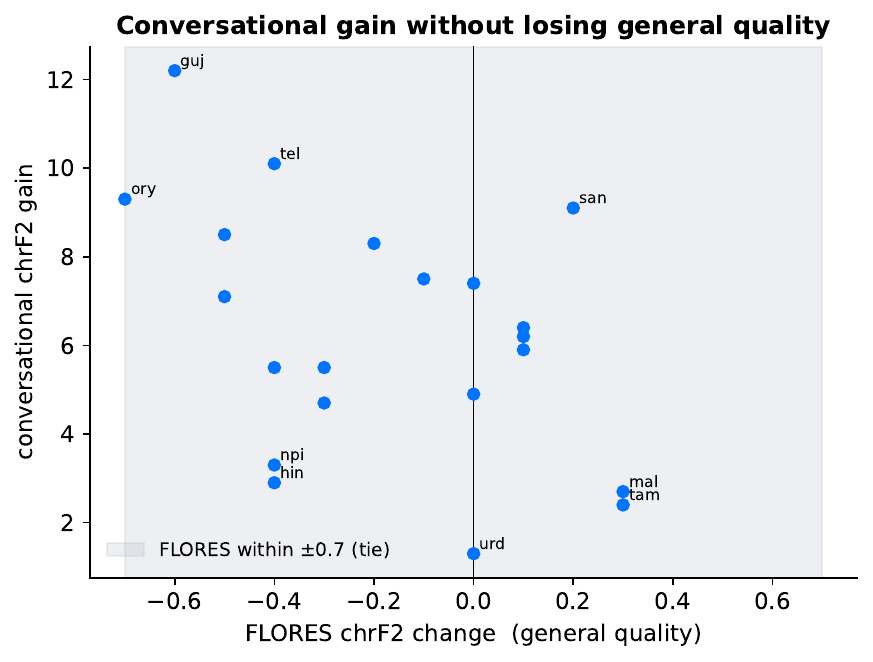}
\caption{Each point is a language. Every language gains on conversation
(all points above zero) while general quality stays inside a 0.7 chrF
band around the base.}
\end{figure}

\hypertarget{significance}{%
\subsubsection{5.3 Significance}\label{significance}}

We run paired bootstrap resampling (sacreBLEU, 1000 resamples, n=500) on
the three languages with the hardest subtitle-based conversational
tests.

\textbf{Table 3: Paired bootstrap, base versus mixed-soup (chrF2).}

\begin{longtable}[]{@{}
  >{\raggedright\arraybackslash}p{(\columnwidth - 4\tabcolsep) * \real{0.3333}}
  >{\raggedright\arraybackslash}p{(\columnwidth - 4\tabcolsep) * \real{0.3333}}
  >{\raggedright\arraybackslash}p{(\columnwidth - 4\tabcolsep) * \real{0.3333}}@{}}
\toprule\noalign{}
\begin{minipage}[b]{\linewidth}\raggedright
Lang
\end{minipage} & \begin{minipage}[b]{\linewidth}\raggedright
conv (p)
\end{minipage} & \begin{minipage}[b]{\linewidth}\raggedright
FLORES (p)
\end{minipage} \\
\midrule\noalign{}
\endhead
\bottomrule\noalign{}
\endlastfoot
Hindi & 51.5 to 54.1 (\textbf{p=0.001}) & 61.9 to 61.9 (p=0.372,
n.s.) \\
Tamil & 50.9 to 55.1 (\textbf{p=0.001}) & 64.7 to 64.6 (p=0.290,
n.s.) \\
Malayalam & 46.9 to 49.5 (\textbf{p=0.004}) & 62.5 to 62.8
(\textbf{p=0.038}, up) \\
\end{longtable}

The conversational gains are significant. FLORES shows no significant
drop (not significant for Hindi and Tamil, and significantly
\emph{better} for Malayalam). This is exactly the result §6 shows to be
misleading if read as a quality claim.

\hypertarget{what-the-metric-does-and-does-not-show}{%
\subsection{6. What the metric does and does not
show}\label{what-the-metric-does-and-does-not-show}}

chrF measures overlap with the reference, not quality, and conversation
has two specific traps we want to name rather than hide.

\textbf{The general result is a tie, not a win.} On FLORES the soup
matches the base (Hindi 61.8 versus 61.7, all 21 languages within 0.7).
So we claim that general quality is preserved, not improved.

\textbf{Part of the conversational gain is reference style matching.}
Our conversational references are subtitles, which are casual. The
adapted model is also more casual (informal pronouns, spoken word
order), so it overlaps the reference more and scores higher whether or
not the meaning is better. On the pairs where base and soup disagree,
the soup is consistently more colloquial, so the conversational gain
reads as ``closer to the casual reference style,'' which overlaps with
but is not the same as ``a better translation.''

\textbf{What survives.} Two claims hold up: the recipe preserves general
chrF (FLORES tie, significance tested), and it produces output
measurably closer to human conversational references (significant on the
hardest tests). Whether that reads as higher quality to a human is a
separate question the metric cannot answer. A blind human and
multi-model LLM check on Hindi did not confirm a perceived gain, which
is why we keep every claim scoped to chrF.

\hypertarget{analysis}{%
\subsection{7. Analysis}\label{analysis}}

\textbf{Forgetting is real and the fix works.} Plain conversational
fine-tuning drops FLORES (3.9 on Hindi). Adding the general anchor
before souping brings FLORES back to base level (mean -0.17), which
confirms replay as the mechanism.

\textbf{Gain size is partly a test-set artifact.} The biggest gains
(Gujarati +12.2, Telugu +10.1, Odia +9.2) are in languages whose
conversational test is only BPCC-H-Daily, short and formulaic and in the
same style as training. The trustworthy gains are on languages with
harder subtitle tests (Hindi +2.9, Tamil +2.4, Malayalam +2.7), which
are exactly the ones §5.3 finds significant. So ranking languages by raw
conversational delta across different test styles would be misleading.

\textbf{A floor on low-resource languages.} Santali (FLORES 30.7),
Sanskrit (36.3), Kashmiri (39.1), and Bodo (45.9) stay low after
adaptation. Their general quality is capped by the base model, which is
capped by how little parallel data exists for them. Conversational
adaptation cannot lift that floor; that would need more general data,
which is not public at scale. This is a data limit, not a method limit.

\hypertarget{conclusion}{%
\subsection{8. Conclusion}\label{conclusion}}

Using replay and model souping, two established techniques, we adapt
IndicTrans2 to conversational register across all 21 Indic languages
while preserving general quality, with significant conversational chrF
gains and no significant general loss. The recipe generalizes across the
family, and its limits are set by the base model on the lowest-resource
languages. By the metric, the result is a single drop-in model that
matches the base on FLORES and is significantly closer to human
conversational references. Whether that closeness is a human-perceived
quality gain is the open question, and we are clear that our current
evidence does not settle it.

\hypertarget{limitations}{%
\subsection{9. Limitations}\label{limitations}}

\begin{itemize}
\tightlist
\item
  \textbf{Automatic metrics only.} All numbers are chrF2.
  Character-n-gram metrics measure overlap, not quality, and can
  overstate adaptation gains when the in-domain references have a
  distinct style {[}Callison-Burch et al., 2006; Mathur et al., 2020;
  Freitag et al., 2021{]}. Our references are subtitles, so part of the
  conversational gain plausibly reflects register matching (§6).
  Confirming a human-perceived gain needs a controlled, multi-annotator
  study; this is the priority next step and is not yet conclusive, so we
  make no human-quality claim.
\item
  \textbf{Single training seed.} One run per setup; we do not yet
  quantify training variance (the significance tests address test-set
  variance only).
\item
  \textbf{Significance on a subset.} Paired bootstrap is reported for 3
  of 21 languages, the hardest-test ones. Extending to all 21 is
  straightforward future work.
\item
  \textbf{Register differs across languages.} The conversational test
  register differs by language (subtitles versus daily utterances),
  which confounds cross-language comparison (§7).
\item
  \textbf{One direction.} English to Indic only. Indic to English and
  Indic to Indic are untested and may forget differently.
\end{itemize}

\hypertarget{reproducibility}{%
\subsection{Reproducibility}\label{reproducibility}}

Code, paper, models, and the evaluation harness are public.

\begin{itemize}
\tightlist
\item
  Code and paper:
  https://github.com/Aditya-PS-05/indictrans2-conversational
\item
  Model:
  https://huggingface.co/adipras1407/indictrans2-en-indic-1B-conversational
\end{itemize}

Key scripts: data (\texttt{src/data/build\_multiling\_conv.py},
\texttt{src/data/build\_ml\_mixed.py}), training
(\texttt{src/train/finetune\_it2\_ml.py}), souping
(\texttt{src/train/make\_soup.py}), evaluation
(\texttt{src/eval/ml\_eval.py}), significance
(\texttt{src/eval/sig\_test.py}). Released models:
\texttt{ft\_it2\_ml\_mixed} and its soup
\texttt{ft\_it2\_ml\_mixed\_soup}. Environment: Python 3.11,
\texttt{transformers==4.40.2}, \texttt{IndicTransToolkit}, sacreBLEU
2.6.

\hypertarget{references}{%
\subsection{References}\label{references}}

\begin{itemize}
\tightlist
\item
  Callison-Burch, Osborne, Koehn. Re-evaluating the Role of BLEU in MT
  Research. EACL 2006.
\item
  Chu, Dabre, Kurohashi. An Empirical Comparison of Domain Adaptation
  Methods for NMT. ACL 2017.
\item
  Freitag, Foster, Grangier, et al.~Experts, Errors, and Context: A
  Large-Scale Study of Human Evaluation for MT. TACL 2021.
\item
  Gala et al.~IndicTrans2: Towards High-Quality and Accessible MT for
  all 22 Scheduled Indian Languages. TMLR 2023.
\item
  Koehn. Statistical Significance Tests for MT Evaluation. EMNLP 2004.
\item
  Lison and Tiedemann. OpenSubtitles2016. LREC 2016.
\item
  Mathur, Baldwin, Cohn. Tangled up in BLEU: Reevaluating the Evaluation
  of Automatic MT Metrics. ACL 2020.
\item
  Nakazawa et al.~Overview of the Workshop on Asian Translation. WAT
  2024.
\item
  NLLB Team (Costa-jussà et al.). No Language Left Behind. arXiv 2022.
  (FLORES-200)
\item
  Popović. chrF: Character n-gram F-score for Automatic MT Evaluation.
  WMT 2015.
\item
  Post. A Call for Clarity in Reporting BLEU Scores (sacreBLEU). WMT
  2018.
\item
  Robins. Catastrophic Forgetting, Rehearsal and Pseudorehearsal.
  Connection Science 1995.
\item
  Wortsman et al.~Model Soups. ICML 2022a.
\item
  Wortsman et al.~Robust Fine-tuning of Zero-shot Models (WiSE-FT). CVPR
  2022b.
\end{itemize}

\end{document}